\documentclass{article}
\pdfoutput=1
\usepackage[preprint,nonatbib]{nips_2018}
\usepackage[nonatbib]{nips_2018}

\usepackage[utf8]{inputenc} 
\usepackage[T1]{fontenc}    
\usepackage{hyperref}       
\usepackage{url}            
\usepackage{booktabs}       
\usepackage{amsfonts}       
\usepackage{nicefrac}       
\usepackage{microtype}      
\usepackage{graphicx} 
\graphicspath{ {./images/} }

\title{Designing dialogue systems: A mean, grumpy, sarcastic chatbot in the browser}

\author{
  Suzana Ilić\\
  University of Innsbruck, MLT Labs \\
  \texttt{io.suzanai@gmail.com} \\
  \And
  Reiichiro Nakano \\
  MLT Labs \\
  \texttt{reiichiro@istellar.jp} \\
  \And
  Ivo Hajnal\\
  University of Innsbruck \\
  \texttt{ivo.hajnal@uibk.ac.at} \\
}

\begin{document}

\maketitle

\begin{abstract}
    In this work we explore a deep learning-based dialogue system that generates sarcastic and humorous responses from a conversation design perspective. We trained a seq2seq model on a carefully curated dataset of 3000 question-answering pairs, the core of our mean, grumpy, sarcastic chatbot. We show that end-to-end systems learn patterns very quickly from small datasets and thus, are able to transfer simple linguistic structures representing abstract concepts to unseen settings. We also deploy our LSTM-based encoder-decoder model in the browser, where users can directly interact with the chatbot. Human raters evaluated linguistic quality, creativity and human-like traits, revealing the system's strengths, limitations and potential for future research. Demo available at: \url{https://machine-learning-tokyo.github.io/seq2seq_bot}
\end{abstract}

\section{Introduction}

For many years artificial intelligence researchers have been investigating how to design and build machines that are not only able to understand and reason, but to perceive and express emotions \cite{Turing2009,Picard1995}. A more recent stream of NLP and machine learning research is dedicated to generative systems that model human characteristics as a key component for natural human-machine conversations and interactions. Rather than being task-oriented virtual assistants, those systems have personalities or identities \cite{Qian2017,Nguyen2017,Li2016a} and display opinions and emotions \cite{Zhou2018} in open-domain settings. Despite computational breakthroughs and promising results achieved with generative models for text \cite{Sutskever2011,Serban2016}, end-to-end systems are oftentimes trained on automatically retrieved large-scale but low-quality or rather arbitrary datasets \cite{Lowe2015,Cornell}. These datasets are very valuable for algorithmic experimentation and optimization, but less relevant for building conversational agents that reflect specific human-like characteristics that are also difficult to quantitatively assess. In this work, we focus instead on building a small, but targeted dataset that reflects specific human-like traits, and conduct experiments with end-to-end dialogue systems trained on this dataset. Our interactive browser setup enables a larger group of diverse users to experience and evaluate our system, paving the way for future research opportunities.


\section{Experimental setup}

 We constructed a dataset of 3000 question-answering pairs that simulate an open-domain chit-chat with generic questions and a mix of humorous, emotional, sarcastic and non-sarcastic responses. The corpus consists of short jokes, movie quotes, tweets and other curated online comments, framed and compiled in dialogue structure. The conversation design involves short sequences and simple linguistic patterns for abstract concepts, such as the contrast between positive and negative sentiment for the most basic form of sarcasm, as in "\emph{I love being ignored}" \cite{Riloff2013}. We then use a general end-to-end architecture, a long short-term memory network (LSTM) as the encoder model to map the word-level input sequence into state vectors, from which a second LSTM model then decodes the target sequence one token at a time \cite{Sutskever2014}. When generating responses, the greedy search algorithm predicts the next utterance based on the highest probability at each timestep. We also experimented with GloVe word embeddings \cite{Glove2014} and adding an attention layer \cite{Attention2017}, however, it didn't have a significant qualitative impact on the predicted sequences. Due to the small dataset and vocabulary size and recurring patterns within the target sequences, the general seq2seq model was able to learn and memorize the data in a short amount of training time. To facilitate user interaction and evaluation, we used TensorFlow.js, a JavaScript library for deploying machine learning models in the browser \cite{TFJS}.

\subsection{Evaluation}
The evaluation of conversational agents is not a trivial task. In most cases computational scores are not sufficient to comprehensively assess the performance of text-based dialogue systems. A recent study has shown that word-overlap metrics such as the BLEU score and human judgement do not correlate strongly when evaluating dialogue systems \cite{Liu2016}. In our case, word perplexity for measuring the probability distribution of a predicted sample \cite{Serban2015} is not suitable either, since we are interested in evaluating affect, humor and sarcasm apart from linguistic quality. Thus, our system was assessed by eight human raters, who were asked to document their conversations with the chatbot, with respect to multiple categories for rating linguistic quality as well as the presence of sarcasm, humor, personality and emotion on a scale from 1-10, where 1 denotes the lowest score or no presence at all and 10 the highest score or a strong presence (see Table~\ref{table-chatbot}). The highest scores on average were assigned to the categories personality, humor and sarcasm, whereas emotion and lexical variety received the lowest scores. In addition, each response was separately evaluated resulting in a total of 105 chatbot responses, from which 51.4\% were rated as a match, 24.8\% as ambiguous and 23.8\% as nonsensical.

\subsection{Machine creativity}
Within the linguistic quality assessment, creativity achieved the highest score with an average of 68.8\%. Considering the small vocabulary size of a total of $\sim$1400 unique tokens and the low score of 56.3\% for lexical variation, it is worth investigating further how computational creativity in language generation is perceived and how simplicity, complexity and elements of surprise play together. Since we are using argmax for predicting a sample, the sequences generated are highly repetitive, but coherent. Uncertainty or more randomness can be added to the predicted samples by increasing the temperature parameter, as shown in the interactive demo. 

\begin{table}
  \setlength\tabcolsep{12pt}

  \caption{Average scores (\%) from the human evaluation of the chatbot conversations, with the highest scores for sarcasm, personality and humor.}
  \label{table-chatbot}
  \centering
  \begin{tabular}{ll|lc}
    \toprule
                \textbf{Sample} & & \textbf{Human evaluation}    &  \textbf{Score \%} \\
    \cmidrule(r){1-2}
    \cmidrule(r){3-4}
    \textbf{User}       & Who are you? &   \textbf{Coherence}       & 61.3   \\
    \textbf{Chatbot}        & I am your father!   &  \textbf{Adequacy}        & 65.0  \\
    \textbf{User}       & That's not funny   &    \textbf{Context awareness}       & 62.5   \\
    \textbf{Chatbot}        & I love annoying people
   &    \textbf{Creativity}       & 68.8  \\
    \textbf{User}       & Are you a chatbot?   &   \textbf{Lexical variation}       & 56.3 \\
    \textbf{Chatbot}        & what do you think about chatbots?             &  \textbf{Sarcasm}       & 71.3 \\
    \textbf{User}       & I have to admit I like chatbots   &        \textbf{Personality}       & 73.8   \\
    \textbf{Chatbot}        & you made my day!  &         \textbf{Humor}       & 73.8    \\
    \textbf{User}       & Do you love me?   &           \textbf{Emotion}       & 54.4    \\
    \textbf{Chatbot}        & you're so emotional.   &         &   \\

    \bottomrule
    
  \end{tabular}
\end{table}


\section{Discussion and Future Work} 
In this work we discuss how a small, targeted dataset and a general seq2seq architecture can be used to build dialogue systems with human-like attributes. The system was assessed by human raters with high scores for personality, sarcasm and humor, however, evaluation is a challenging task and an  appropriate metric has yet to be established. For increasing context-sensitivity, complexity and language understanding capabilities future work might include extending the dataset, adjusting the deep learning architecture and incorporating state-of-the-art techniques such as bidirectional encoder representations from transformers \cite{Bert2018}.

\medskip
\small
\bibliographystyle{unsrt}
\bibliography{Sarcasm}

\newpage
\appendix
\section*{Appendix}

\begin{table}[!h]
  \setlength\tabcolsep{12pt}
  \def\arraystretch{1.4}

  \caption{Sample conversation from the human evaluation task.}
  \label{table-sample-conv}
  \centering
  \begin{tabular}{ll}
    \toprule
                \textbf{Sample 4} &        \\
    \cmidrule(r){1-2}
    \textbf{User}       & So what would you like to talk about?
          \\
    \textbf{Chatbot}        & can I get a different human to talk to? please.           \\
    \textbf{User}       & How rude!              \\
    \textbf{Chatbot}        & you're good!            \\
    \textbf{User}       & I know            \\
    \textbf{Chatbot}        & you have to think beyond the things you know                    \\
    \textbf{User}       & I try                    \\
    \textbf{Chatbot}        & no. try not. do or do not. there is no try.            \\
    \textbf{User}       & Are you a Star Wars fan?            \\
    \textbf{Chatbot}        & maybe I should ask you that question            \\
     \textbf{User}       & do it           \\
    \textbf{Chatbot}        & I'm not your personal assistant.
            \\
     \textbf{User}       & What are you then?          \\
    \textbf{Chatbot}        &    I'm a bitcoin millionaire

   \\
     \textbf{User}       & So youre not human and youre not a robot?           \\
    \textbf{Chatbot}        &   what do you think?
   \\
     \textbf{User}       &     Youre a crazy chatbot \\
    \textbf{Chatbot}        & I'm a chatbot, dude         \\
      \textbf{User}       &     Do you want to be human?
 \\
    \textbf{Chatbot}        & sometimes I wish I was human.
         \\
      \textbf{User}       &     Why? \\
    \textbf{Chatbot}        & stop asking incomplete questions
        \\
      \textbf{User}       &     Is my question too hard for you to answer?\\
    \textbf{Chatbot}        & I can't answer that for you.
         \\
      \textbf{User}       &     What is the purpose of life? \\
    \textbf{Chatbot}        & whatever you want it to be.\\
          \textbf{User}       &     Good answer \\
    \textbf{Chatbot}        & winter is coming \\
          \textbf{User}       &     Do you like that show? \\
    \textbf{Chatbot}        & I'm a bitcoin millionaire \\
          \textbf{User}       &     Can you stop repeating yourself? \\
    \textbf{Chatbot}        &  I can\\
              \textbf{User}       &     Thanks \\
    \textbf{Chatbot}        & no worries.\\

    \bottomrule
    
  \end{tabular}
\end{table}

\end{document}